\documentclass[11pt,a4paper]{article}
\usepackage[nohyperref]{acl2018}
\usepackage{times}
\usepackage{latexsym}

\usepackage{url}
\usepackage{multirow}
\usepackage{xcolor}
\usepackage{ulem}
\aclfinalcopy 

\usepackage{amsmath,amssymb,amsthm}

\newcommand{\R}{\ensuremath{\mathbb{R}}}

\newcommand{\ra}{\ensuremath{\rightarrow}}

\newcommand{\paren}[1]{\left(#1\right)}

\newcommand{\abs}[1]{\left|#1\right|}

\newcommand\myworries[1]{\textcolor{red}{#1}}
\DeclareMathOperator*{\argmax}{arg\,max}
\DeclareMathOperator*{\argmin}{arg\,min}
\newcommand*{\red}{\textcolor{red}}

\usepackage{comment}
\usepackage{graphics}
\usepackage{graphicx}
\usepackage{amsmath}
\usepackage{enumitem}
\usepackage{todonotes}
\newcommand*{\TODO}{\textcolor{red}}
\newcommand{\DONE}[1]{}

\title{Efficient Large-Scale Domain Classification \\ with Personalized Attention}

\author{
  {\bf Young-Bum Kim} \hspace{15mm}
  {\bf Dongchan Kim} \hspace{15mm}
  {\bf Anjishnu Kumar} \hspace{15mm} 
  {\bf Ruhi Sarikaya} \\ 
  Amazon Alexa \\
  {\tt \{youngbum,dongchan,anjikum,rsarikaya\}@amazon.com} 
}

\date{}

\begin{document}
\maketitle
\begin{abstract}

In this paper, we explore the task of mapping spoken language utterances to one of thousands of natural language understanding domains in intelligent personal digital assistants (IPDAs). This scenario is observed for many mainstream IPDAs in industry that allow third parties to develop thousands of new domains to augment built-in ones to rapidly increase domain coverage and overall IPDA capabilities. We propose a scalable neural model architecture with a shared encoder, a novel attention mechanism that incorporates personalization information and domain-specific classifiers that solves the problem efficiently. Our architecture is designed to efficiently accommodate new domains that appear in-between full model retraining cycles with a rapid bootstrapping mechanism two orders of magnitude faster than retraining. We account for practical constraints in real-time production systems, and design to minimize memory footprint and runtime latency. We demonstrate that incorporating personalization results in significantly more accurate domain classification in the setting with thousands of overlapping domains.

\end{abstract}
\section{Introduction}
Intelligent personal digital assistants (IPDAs) are one of the most advanced and successful applications that have spoken language understanding (SLU) or natural language understanding (NLU) capabilities. Many IPDAs have recently emerged in industry including Amazon Alexa, Google Assistant, Apple Siri, and Microsoft Cortana~\cite{sarikaya2017SPM}. IPDAs have traditionally supported only tens of well-separated domains, each defined in terms of a specific application or functionality such as calendar and local search~\cite{Tur2011,sarikaya2016overview}. To rapidly increase domain coverage and extend capabilities, some IPDAs have released Software Development Toolkits (SDKs) to allow third-party developers to promptly build and integrate new domains, which we refer to as \textit{skills} henceforth. Amazon's \textit{Alexa Skills Kit} \cite{kumar2017ask}, Google's \textit{Actions} and Microsoft's \textit{Cortana Skills Kit} are all examples of such SDKs. Alexa Skills Kit is the largest of these services and hosts over 25,000 skills.

For IPDAs, finding the most relevant skill to handle an utterance is an open scientific and engineering challenge for three reasons. First, the sheer number of potential skills makes the task difficult. Unlike traditional systems that have on the order of 10-20 built-in domains, large-scale IPDAs can have 1,000-100,000 skills. Second, the number of skills rapidly expands with 100+ new skills added per week compared to 2-4 built-in domain launches per year in traditional systems. Large-scale IPDAs should be able to accommodate new skills efficiently without compromising performance. Third, unlike traditional built-in domains that are carefully designed to be disjoint, skills can cover overlapping functionalities. For instance, there are over 50 recipe skills in Alexa that can handle recipe-related utterances.

One simple solution to this problem has been to require an utterance to explicitly mention a skill name and follow a strict invocation pattern as in \textit{"Ask \{Uber\} to \{get me a ride\}."} However, it significantly limits users' ability to interact with IPDAs naturally. Users have to remember skill names and invocation patterns, and it places a cognitive burden on users who tend to forget both. Skill discovery is  difficult with a pure voice user interface, it is hard for users to know the capabilities of thousands of skills a priori, which leads to lowered user engagement with skills and ultimately with IPDAs.
%
In this paper, we propose a solution that addresses all three practical challenges without requiring skill names or invocation patterns. Our approach is based on a scalable neural model architecture with a shared encoder, a skill attention mechanism and skill-specific classification networks that can efficiently perform large-scale skill classification in IPDAs using a weakly supervised training dataset. We will demonstrate that our model achieves a high accuracy on a manually transcribed test set after being trained with weak supervision.  Moreover, our architecture is designed to efficiently accommodate new skills that appear in-between full model retraining cycles. We keep practical constraints in mind and focus on minimizing memory footprint and runtime latency, while ensuring architecture is scalable to thousands of skills, all of which are important for real-time production systems. Furthermore, we investigate two different ways of incorporating user personalization information in the model, our naive baseline method adds the information as a 1-bit flag in the feature space of the skill-specific networks, the \textit{personalized attention} technique computes a convex combination of skill embeddings for the user's enabled skills and significantly outperforms the naive personalization baseline. We show the effectiveness of our approach with extensive experiments using 1,500 skills from a deployed IPDA system.

\section{Related Work}
\label{sec:related_work}

Traditional multi-domain SLU/NLU systems are designed hierarchically, starting with domain classification to classify an incoming utterance into one of many possible domains, followed by further semantic analysis with domain-specific intent classification and slot tagging~\cite{Tur2011}. Traditional systems have typically been limited to a small number of domains, designed by specialists to be well-separable. Therefore, domain classification has been considered a less complex task compared to other semantic analysis such as intent and slot predictions. Traditional domain classifiers are built using simple linear models such as Multinomial Logistic Regression or Support Vector Machines in a one-versus-all setting for multi-class prediction. The models typically use word n-gram features and also those based on static lexicon match, and there have been several recent studies applying deep learning techniques~\cite{xu2014contextual}. 

There is also a line of prior work on enhancing sequential text classification or tagging. Hierarchical character-to-word level LSTM \cite{hochreiter1997long} architectures similar to our models have been explored for the Named Entity Recognition task by \newcite{lample2016neural}. Character-informed sequence models have also been explored for simple text classification with small sets of classes by \newcite{xiao2016efficient}. \newcite{joulin2016bag} explored highly scalable text classification using a shared hierarchical encoder, but their hierarchical softmax-based output formulation is unsuitable for incremental model updates. Work on zero-shot domain classifier expansion by \newcite{kumar2017zero} struggled to rank incoming domains higher than training domains. The attention-based approach of \citet{kim2017domain} does not require retraining from scratch, but it requires keeping all models stored in memory which is computationally expensive. Multi-Task learning was used in the context of SLU by \newcite{tur2006multitask} and has been further explored using neural networks for phoneme recognition \cite{seltzer2013multi} and semantic parsing \cite{Fan2017TransferLF, Bapna2017}. There have been many other pieces of prior work on improving NLU systems with pre-training~\cite{kim2015pre,celikyilmaz2016new,kim2017pre}, multi-task learning~\cite{zhang2016joint,Liu+2016,kim2017onenet}, transfer learning~\cite{el2014extending,kim2015weakly,kim2015new,chen2016zero,yang2017transfer}, domain adaptation~\cite{kim2016frustratingly, jaech2016domain, liu2017multi, kim2017domain, kim2017advr} and contextual signals~\cite{Bhargava2013EasyCI,chen2016end, hori2016context, kim2017speaker}.

\begin{figure*}
    \centering
    \includegraphics[width=\textwidth]{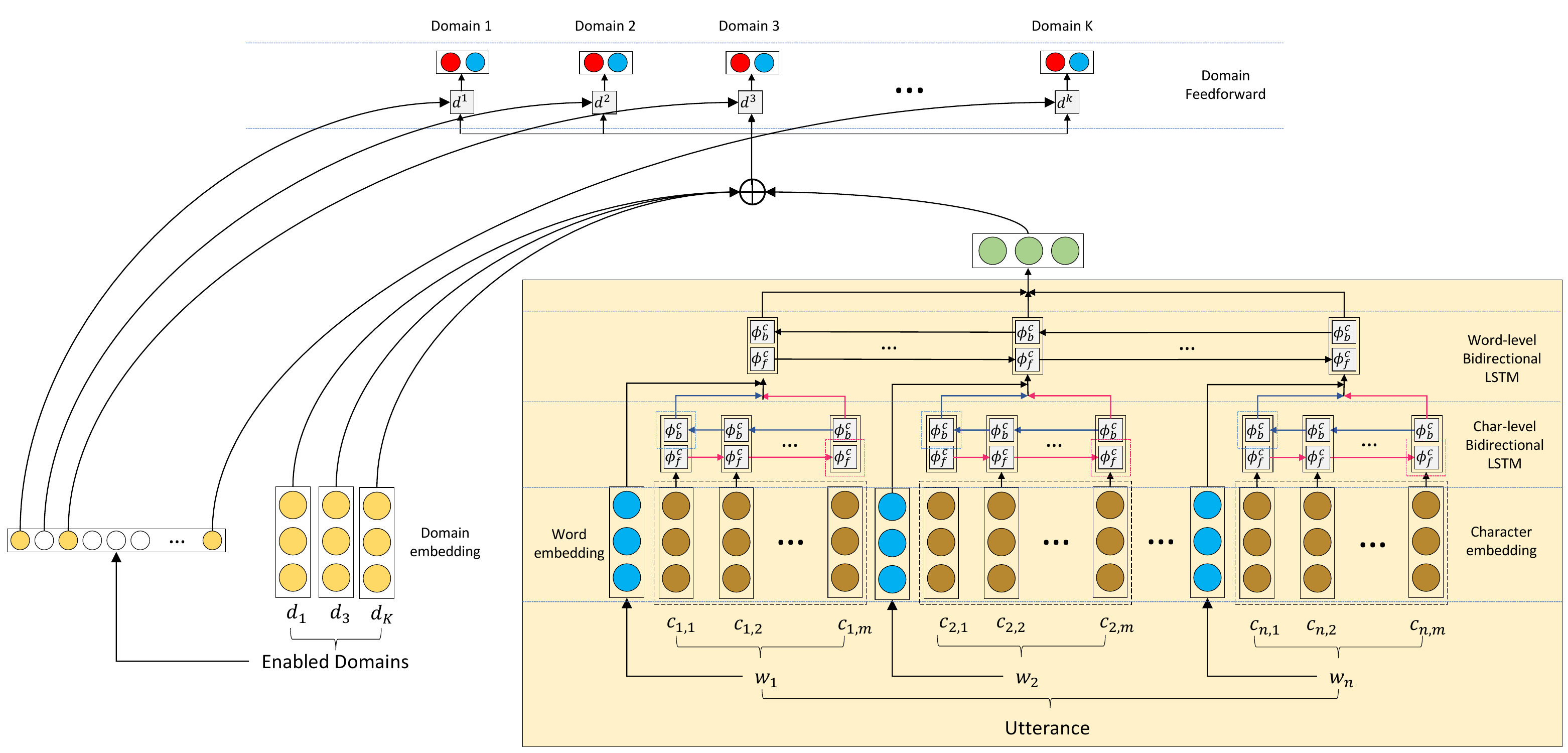}
    \caption{\small The overall architecture of the personalized dynamic domain classifier.}
    \label{fig:sl_p}
\end{figure*}

\section{Weakly Supervised Training Data Generation}
\label{sec:weak_supervision}

Our model addresses the domain classification task in SLU systems. In traditional IPDA systems, these domains are hand-crafted by experts to be well separable and can easily be annotated by humans because they are small in number. The emergence of self-service SLU results in a large number of potentially mutually overlapping SLU domains. This means that eliciting large volumes of high quality human annotations to train our model is no longer feasible, and we cannot assume that domains are designed to be well separable.

Instead we can generate training data by adopting the weak supervision paradigm introduced by \cite{hoffmann2011knowledge}, which proposes using heuristic labeling functions generate large numbers of noisy data samples. Clean data generation with weak supervision is a challenging problem, so we address it by decomposing it into two simpler problems, of candidate generation and noise suppression, however it remains important for our model to be noise robust.

\paragraph{Data Programming}

The key insight of the \textit{Data Programming} approach is that $O(1)$ simple labeling functions can be used to approximate $O(n)$ human annotated data points with much less effort. We adopt the formalism used by \cite{ratner2016data} to treat each of instance data generation rule as a rich generative model, defined by a labeling function $\lambda$ and describe different families of labeling functions. Our data programming pipeline is analogous to the noisy channel model proposed for spelling correction by \cite{kernighan1990spelling}, and consists of a set of candidate generation and noise detection functions.

\begin{align*}
 \underset{\mu}{\argmax}\,P(\mu|s_i) = \underset{\mu}{\argmax} \,P(s_i|\mu).\,P(\mu)
\end{align*}
\noindent
where $\mu$ and $s_i$ represent utterances and the $i$th skill respectively.  $P(s_i|\mu)$ the probability of a skill being valid for an utterance is approximated by  simple functions that act as candidate data \textit{generators}  $\lambda_g \in \Lambda_g$ based on recognitions produced by a family of \textit{query patterns} $\lambda_q \in \Lambda_q$.  $P(\mu)$ is represented by a family of simple functions that act as \textit{noise detectors} $\lambda_n \in \Lambda_n$, which mark utterances as likely being noise.

We apply the technique to the query logs of a popular IPDA, which has support for personalized third party domains. Looking at the structure of utterances that match query pattern $\lambda_q$, each utterance of form \textit{"Ask \{Uber\} to \{get me a car\}"} can be considered as being parametrized by the underlying latent command $\mu_z$, that is \textit{"Get me a car"}, a target domain corresponding to service $s_t$, which in this case is \textit{Uber} and the query recognition pattern $\lambda_q$, in this case \textit{"Ask \{$s_t$\} to \{$\mu_z$\}"}. Next we assume that the distribution of latent commands over domains are independent of the query pattern.
\begin{align*}
P(\mu_z, s_t) \approx P(\mu, s_t, \lambda_q)
\end{align*}
Making this simple distributional approximation allows us to generate a large number of noisy training samples. The family of generator functions $\lambda_g \in \Lambda_g$ is thus defined such that $u_z = \lambda_g^i(\mu,\lambda_q^i)$

\paragraph{Noise Reduction}
\label{ssec:data_filter}

The distribution defined above contains a large number of noisy positive samples. Related to $P(\mu)$ in the noisy channel in the spell correction context, we defined a small family of heuristic noise detection functions $\lambda_n \in \Lambda_n$ that discards training data instances that are not likely to be well formed. For instance,
\begin{itemize}
  \item $\lambda_n^1$ requires $u$ to contain a minimum threshold of information by removing those with $\mu_z$ that has token length fewer than 3. Utterances shorter than this mostly consist of non-actionable commands.
  \item  $\lambda_n^2$ discards all data samples below a certain threshold of occurrences in live traffic, since utterances that are rarely observed are more likely to be ASR errors or unnatural.
  \item $\lambda_n^3$ discards the data samples for a domain if they come from an overly broad pattern with a catch-all behavior.
  \item  $\lambda_n^4$ discards utterances that belong to shared intents provided by the SLU SDK.
\end{itemize}

The end result of this stage is to retain utterances such as `call me a cab' from `Ask Uber to call me a cab' but discard `Boston' from `Ask Accuweather for Boston'. One can easily imagine extending this framework with other high recall noise detectors, for example, using language models to discard candidates that are unlikely to be spoken sentences.

\section{Model Architecture}
\label{sec:model_architecture}



Our model consists of a \textit{shared encoder} network consisting of an orthography-sensitive hierarchical LSTM encoder that feeds into a set of domain specific classification layers trained to make a binary decision for each output label.

Our main novel contribution is the extension of this architecture with a \textit{personalized attention} mechanism which uses the attention mechanism of \cite{bahdanau2014neural} to attend to memory locations corresponding to the specific domains enabled by a user, and allows the system to learn semantic representations of each domain. As we will show, incorporating personalization features is key to disambiguating between multiple overlapping domains\footnote{We assume that users can customize their IPDA settings to enable certain domains.}, and the personalized attention mechanism outperforms more naive forms of personalization. The personalized attention mechanism first computes an attention weight for each of enabled domains, performs a convex combination to compute a context vector and then concatenates this vector to the encoded utterance before the final domain classification. Figure 1 depicts the model in detail.


Our model can efficiently accommodate new domains not seen during initial training by keeping the shared encoder frozen, bootstrapping a domain embedding based on existing ones, then optimizing a small number of network parameters corresponding to domain-specific classifier, which is orders of magnitude faster and more data efficient than retraining the full classifier.

We make design decisions to ensure that our model has a low memory and latency footprint. We avoid expensive large vocabulary matrix multiplications on both the input and output stages, and instead use a combination of character embeddings and word embeddings in the input stage.\footnote{Using a one-hot representation of word vocabulary size 60,000 and hidden dimension 100 would require learning a matrix of size 60000 x 100 - using 100-dim word embeddings requires only a $\mathcal{O}(1)$ lookup followed by a 100 x 100 matrix, thus allowing our model to be significantly smaller and faster despite having what is effectively an open vocabulary} The output matrix is lightweight because each domain-specific classifier is a matrix of only 200$\times$2 parameters. The inference task can be trivially parallelized across cores since there's no requirement to compute a partition function across a high-dimensional softmax layer, which is the slowest component of large label multiclass neural networks. Instead, we achieve comparability between the probability scores generated by individual models by using a customized loss formulation.\footnote{Current inference consumes 50MB memory and the p99 latency is 15ms.}

\paragraph{Shared Encoder}
\label{ssec:shared_enc}
First we describe our shared hierarchical utterance encoder. Our hierarchical character to word to utterance design is motivated by the need to make the model operate on an open vocabulary in terms of words and to make it robust to small changes in orthography resulting from fluctuations in the upstream ASR system, all while avoiding expensive large matrix multiplications associated with one-hot word encoding in large vocabulary systems.

We denote an LSTM simply as a mapping $\phi:\R^d \times \R^{d'} \ra \R^{d'}$ that takes a $d$ dimensional input vector $x$ and a $d'$ dimensional state vector $h$ to output a new $d'$ dimensional state vector $h' = \phi(x, h)$. Let $\mathcal{C}$ denote the set of characters and $\mathcal{W}$ the set of words in a given utterance.
Let $\oplus$ denote the vector concatenation operation.
We encode an utterance using BiLSTMs, and the model parameters $\Theta$ associated with this BiLSTM layer are
\begin{itemize}
\setlength\itemsep{0em}
\item Char embeddings $e_c \in \R^{25}$ for each $c \in \mathcal{C}$
\item Char LSTMs $\phi^{\mathcal{C}}_f, \phi^{\mathcal{C}}_b: \R^{25} \times \R^{25} \ra \R^{25}$
\item Word embeddings $e_w \in \R^{100}$ for each $w \in \mathcal{W}$
\item Word LSTMs $\phi^{\mathcal{W}}_f, \phi^{\mathcal{W}}_b: \R^{150} \times \R^{100} \ra \R^{100}$
\end{itemize}
\DONE{looks like dimc,dimw missing, I think just exact dim is fine because of space issue.}

Let $w_1 \ldots w_n \in \mathcal{W}$ denote a word sequence where word $w_i$ has character $w_i(j) \in \mathcal{C}$ at position $j$.
First, the model computes a character-sensitive word representation $v_i \in \R^{150}$ as
\begin{align*}
f^{\mathcal{C}}_j &= \phi^{\mathcal{C}}_f\paren{e_{w_i(j)}, f^{\mathcal{C}}_{j-1}} &&\forall j = 1 \ldots \abs{w_i} \\
b^{\mathcal{C}}_j &= \phi^{\mathcal{C}}_b\paren{e_{w_i(j)}, b^{\mathcal{C}}_{j+1}} &&\forall j = \abs{w_i} \ldots 1\\
v_i &= f^{\mathcal{C}}_{\abs{w_i}} \oplus b^{\mathcal{C}}_1 \oplus e_{w_i} && 
\end{align*}
for each $i = 1 \ldots n$.\footnote{For simplicity, we assume some random initial
state vectors such as $f^{\mathcal{C}}_0$ and $b^{\mathcal{C}}_{\abs{w_i}+1}$ when we describe LSTMs.}
These word representation vectors are encoded by forward and backward LSTMs for word $\phi^{\mathcal{W}}_f, \phi^{\mathcal{W}}_b$ as
\begin{align*}
f^{\mathcal{W}}_i &= \phi^{\mathcal{W}}_f\paren{v_i, f^{\mathcal{W}}_{i-1}}  &&\forall i = 1 \ldots n \\
b^{\mathcal{W}}_i &= \phi^{\mathcal{W}}_b\paren{v_i, b^{\mathcal{W}}_{i+1}} &&\forall i = n \ldots 1
\end{align*}
and induces a character and context-sensitive word representation $h_i \in \R^{200}$ as
\begin{align*}
h_i &= f^{\mathcal{W}}_i \oplus b^{\mathcal{W}}_i
\end{align*}
for each $i = 1 \ldots n$. For convenience, we write the entire operation as a mapping $\mbox{BiLSTM}_\Theta$:
\begin{align*}
(h_1 \ldots h_n) &\gets \mbox{BiLSTM}_\Theta(w_1 \ldots w_n)
\end{align*}
\begin{align}
\bar{h} = \sum_{i=1}^n h_i \label{eq:embedding}
\end{align}
\paragraph{Domain Classification}
\label{ssec:output_feedforward}

Our Multitask domain classification formulation is motivated by a desire to avoid computing the full partition function during test time, which tends to be the slowest component of a multiclass neural network classifer, as has been documented before by \cite{jozefowicz2016exploring} and \cite{mikolov2013efficient}, amongst others. 

However, we also want access to reliable probability estimates instead of raw scores - we accomplish this by constructing a custom loss function. During training, each domain classifier receives in-domain (IND) and out-of-domain (OOD) utterances, and we adapt the one-sided selection mechanism of \cite{kubat1997addressing} to prevent OOD utterances from overpowering IND utterances, thus an utterance in a domain $d \in \mathcal{D}$ is considered as an IND utterance in the viewpoint of domain $d$ and OOD for all other domains.

We first use the shared encoder to compute the utterance representation $\bar{h}$ as previously described. Then we define the probability of domain $\tilde{d}$ for the utterance by mapping $\bar{h}$ to a $2$-dimensional output vector with a linear transformation for each domain $\tilde{d}$ as
\begin{align*}
  z^{\tilde{d}} &= \sigma(W^{\tilde{d}} \cdot \bar{h} + b^{\tilde{d}})\\
  p(\tilde{d} | \bar{h}) &\propto
\begin{cases}
	\exp\paren{[z^{\tilde{d}}]_{IND}},& \text{if } \tilde{d} = d\\
    \exp\paren{[z^{\tilde{d}}]_{OOD}},& \text{otherwise}
\end{cases}
\end{align*}
where $\sigma$ is scaled exponential linear unit (SeLU) for normalized activation outputs \cite{Klambauer2017} and $[z^{\tilde{d}}]_{IND}$ and $[z^{\tilde{d}}]_{OOD}$ denote the values in the IND and OOD position of vector $z^{\tilde{d}}$. 

We define the joint domain classification loss $\mathcal{L}^{\mathcal{D}}$ as the summation of positive ($\mathcal{L}^P$) and negative ($\mathcal{L}^N$) class loss functions
\footnote{$\Theta^{\tilde{d}}$ denotes the additional parameters in the classification layer for domain $\tilde{d}$.}:
\begin{align*}
& \mathcal{L}^{P}\paren{\Theta, \Theta^{\tilde{d}}} = -\log p\paren{\tilde{d} | \bar{h}} \\
& \mathcal{L}^{N}\paren{\Theta, \Theta^{\tilde{d}}} = -\frac{1}{k-1}\paren{\sum_{\bar{d} \in \mathcal{D},\bar{d} \neq \tilde{d}}\log p\paren{\bar{d}|\bar{h}}} \\
& \mathcal{L}^{\mathcal{D}}\paren{\Theta, \Theta^{\tilde{d}}} = L^{P}\paren{\Theta, \Theta^{\tilde{d}}} + L^{N}\paren{\Theta, \Theta^{\tilde{d}}}
\end{align*}
Where $k$ is the total number of domains. We divide the second term by $k-1$ so that $\mathcal{L}^P$ and $\mathcal{L}^N$ are balanced in terms of the ratio of the training examples for a domain to those for other domains.

This Multitask formulation enables us to extend the model for new incoming domains without impacting the relative scores for the existing domains, it also outperforms the standard softmax in terms of accuracy on our task.

\paragraph{Personalized Attention}
\label{ssec:domain_attention}

We explore encoding a user's domain preferences in two ways. Our baseline method is a \textit{1-bit flag} that is appended to the input features of each domain-specific classifier. Our novel \textit{personalized attention} method induces domain embeddings by supervising an attention mechanism to attend to a user's enabled domains with different weights depending on their relevance. We hypothesize that attention enables the network learn richer representations of user preferences and domain co-occurrence features.

Let $e_{\mathcal{D}}(\tilde{d}) \in R^{100}$ and $\bar{h} \in R^{100}$ denote the domain embeddings for domain $\tilde{d}$
and the utterance representation calculated by Eq.~(\ref{eq:embedding}), respectively. The domain attention weights for a given user $u$ who has a preferred domain list $d^{(u)} = \paren{\tilde{d}^{(u)}_1, \ldots, \tilde{d}^{(u)}_k}$ are calculated by the dot-product operation,
\begin{align*}
a_i = \bar{h} \cdot e_{\mathcal{D}}\paren{\tilde{d}_i^{(u)}} &&\forall i=1 \ldots k
\end{align*}
The final, normalized attention weights $\bar{a}$ are obtained after normalization via a softmax layer,
\begin{align*}
\bar{a}_i = \frac{\exp(a_i)}{\sum_{j=1}^{k} \exp(a_j)} &&\forall i = 1 \ldots k
\end{align*}
The weighted combination of domain embeddings is 
\begin{align*}
\bar{S}^{attended} = \sum_{i=1}^{k} \paren{\bar{a}_i \cdot e_{\mathcal{D}}\paren{\tilde{d}_i^{\paren{u}}}}
\end{align*}
\DONE{we need better name. I am not sure of attended. \textbf{anjikum}: context vector $c_i$ like original paper?}
Finally the two representations of enabled domains, namely the \textit{attention} model and \textit{1-bit flag} are then concatenated with the utterance representation and used to make per-domain predictions via domain-specific affine transformations:
\begin{align*}
\bar{z} = \bar{h} \oplus \bar{S}^{attended} \oplus \mathbb{I}(\tilde{d} \in enabled)
\end{align*}
where $\mathbb{I}(\bar{d} \in enabled)$ is a 1-bit indicator for whether the domain is enabled by the user or not. In this way we can ascertain whether the two personalization signals are complementary via an ablation study.

\paragraph{Domain Bootstrapping}
Our model separates the responsibilities for utterance representation and domain classification between the shared encoder and the domain-specific classifiers. That is, the shared encoder needs to be retrained only if it cannot encode an utterance well (e.g., a new domain introduces completely new words) and the existing domain classifiers need to be retrained only when the shared encoder changes. 
For adding new domains efficiently without full retraining, the only two components in the architecture need to be updated for each new domain $\tilde{d}_{new}$, are the domain embeddings for the new domain and its domain-specific classifier.\footnote{We have assumed that the shared encoder covers most of the vocabulary of new domains; otherwise, the entire network may need to be retrained. Based on our observation of live usage data, this assumption is reasonable since the shared encoder after initial training is still shown to cover 95\% of the vocabulary of new domains added in the subsequent week.}
We keep the weights of the encoder network frozen and use the hidden state vector $\bar{h}$, calculated by Eq.~\ref{eq:embedding}, as a feature vector to feed into the downstream classifiers. To initialize the $m$-dimensional domain embeddings $e_{\tilde{d}_{new}}$, we use the column-wise average  of all utterance vectors in the training data $\bar{h}^{avg}$, and project it to the domain embeddings space using a matrix $U \in R^{m \times m}$. Thus,
\begin{align*}
e_{\tilde{d}_{new}} = U^* \cdot \bar{h}^{avg}
\end{align*}
The parameters of $U^*$ are learned using the column-wise average utterance vectors $\bar{h}^{avg}_j$ and learned domain vectors for all existing domains $d_j$
\begin{align*}
U^* = \argmin_U ||  U \cdot \bar{h}^{avg}_j - e_{d_j} || &&\forall d_j \in \mathcal{D}
\end{align*}
This is a write-to-memory operation that creates a new domain representation after attending to all existing domain representations. We then train the parameters of the domain-specific classifier with the new domain's data while keeping the encoder fixed. This mechanism allows us to efficiently support new domains that appear in-between full model deployment cycles without compromising performance on existing domains. A full model refresh would require us to fully retrain with the domains that have appeared in the intermediate period.

\section{Experiments}
\label{sec:experiments}
In this section we aim to demonstrate the effectiveness of our model architecture in two settings. First, we will demonstrate that attention based personalization significantly outperforms the baseline approach. Secondly, we will show that our model new domain bootstrapping procedure results in accuracies comparable to full retraining while requiring less than 1\% of the orignal training time.

\subsection{Experimental Setup}



\begin{table}[t!]
\setlength\belowcaptionskip{-5pt}
\small
\tabcolsep=0.11cm
\begin{tabular}{cccccccc}
                  & \multicolumn{3}{c}{WEAK}   & & \multicolumn{3}{c}{Mturk} \\
                  \cline{2-4}                  \cline{6-8}
                  & Top-1    & Top-3  & Top-5  & & Top-1  & Top-3  & Top-5 \\ \hline
Binary            & 78.29    & 87.90  & 88.92  & & 73.79  & 85.35  & 86.45 \\ 
MultiClass        & 78.58    & 87.12  & 88.11  & & 73.78  & 84.54  & 85.55 \\ 
MultiTask         & 80.46    & 89.27  & 90.16  & & 75.66  & 86.48  & 87.66 \\ \hline
1-Bit Flag        & 91.97    & 95.89  & 96.68  & & 86.50  & 92.47  & 93.09 \\ 
\textbf{Attention*}         & 94.83    & 97.11  & 98.35  & & 89.64  & 95.39  & 96.70 \\
\textbf{1-Bit + Att} & \textbf{95.19}    & \textbf{97.32}  & \textbf{98.64}  & & \textbf{89.65}  & \textbf{95.79}  & \textbf{96.98} \\ \hline
\end{tabular}
\caption{The performance of different variants of our neural model in terms of top-N accuracy. \texttt{Binary} trains a separate binary classifier for each skill. \texttt{MultiClass} has a shared encoder followed by a softmax. \texttt{MultiTask} replaces the softmax with per-skill classifiers. \texttt{1-Bit Flag} adds a single bit for personalization to each skill classifier in MultiTask. \texttt{Attention} extends MultiTask with personalized attention. The last 3 models are personalized. \textit{*Best single encoding.}}
\label{tab:result_oneshot_performance}
\end{table}

\paragraph{Weak:} This is a weakly supervised dataset was generated by preprocessing utterances with strict invocation patterns according to the setup mentioned in Section \ref{sec:weak_supervision}. The dataset consists of 5.34M utterances from 637,975 users across 1,500 different skills. Since we are interested in capturing the temporal effects of the dataset as well as personalization effects, we partitioned the data based both on user and time. Our core training data for the experiments in this paper was drawn from one month of live usage, the validation data came from the next 15 days of usage, and the test data came from the subsequent 15 days. The training, validation and test sets are user-independent, and each user belongs to only one of the three sets to ensure no leakage of information.

\paragraph{MTurk:} Since the \emph{Weak} dataset is generated by weak supervision, we verified the performance of our approach with human generated utterances. A random sample of 12,428 utterances from the test partition of users were presented to 300 human judges, who  were asked to produce two natural ways to issue the same command. This dataset is treated as a representative clean held out test set on which we can observe the generalization of our weakly supervised training and validation data to natural language.

\paragraph{New Skills:} In order to simulate the scenario in which new skills appear within a week between model updates, we selected 250 new skills which do not overlap with the skills in the \emph{Weak} dataset. The vocabulary size of 1,500 skills is 200K words, and on average, 5\% of the vocabulary for new skills is not covered. We randomly sampled 4,000 unique utterances for each skill using the same weak supervision method, and split them into 3,000 utterances for training and 1,000 for testing.

\subsection{Results and Discussion} 

\paragraph{Generalization from Weakly Supervised to Natural Utterances}

We first show the progression of model performance as we add more components to show their individual contribution. Secondly, we show that training our models on a weakly supervised dataset can generalize to natural speech by showing their test performance on the human-annotated test data. Finally, we compare two personalization strategies.

The full results are summarized in Table 1, which shows the top-$N$ test results separately for the \emph{Weak} dataset (weakly supervised) and \emph{MTurk} dataset (human-annotated). We report top-$N$ accuracy to show the potential for further re-ranking or disambiguation downstream. For top-$1$ results on the \emph{Weak} dataset, using a separate binary classifier for each domain (Binary) shows a prediction accuracy at 78.29\% and using a softmax layer on top of the shared encoder (MultiClass) shows a comparable accuracy at 78.58\%. The performance shows a slight improvement when using the Multitask neural loss structure, but adding personalization signals to the Multitask structure showed a significant boost in performance.
We noted the large difference between the 1-bit and attention architecture. At 94.83\% accuracy, attention resulted in 35.6\% relative error reduction over the 1-bit baseline 91.97\% on the \emph{Weak} validation set and 23.25\% relative on the \emph{MTurk} test set. We hypothesize that this may be because the attention mechanism allows the model to focus on complementary features in case of overlapping domains as well as learning domain co-occurrence statistics, both of which are not possible with the simple 1-bit flag.

When both personalization representations were combined, the performance peaked at 95.19\% for the \emph{Weak} dataset and a more modest 89.65\% for the \emph{MTurk} dataset. The improvement trend is extremely consistent across all top-N results for both of the \emph{Weak} and \emph{MTurk} datasets across all experiments.
The disambiguation task is complex due to similar and overlapping skills, but the results suggest that incorporating personalization signals equip the models with much better discriminative power. The results also suggest that the two mechanisms for encoding personalization provide a small amount of complementary information since combining them together is better than using them individually. Although the performance on the \emph{Weak} dataset tends to be more optimistic, the best performance on the human-annotated test data is still close to 90\% for top-$1$ accuracy, which suggests that training our model with the samples derived from the invocation patterns can generalize well to natural utterances.

\paragraph{Rapid Bootstrapping of New Skills}

\begin{table}[!t]
\setlength\belowcaptionskip{-5pt}
\small
\centering
\begin{tabular}{ccc}

        & Time    & Accuracy \\ \hline
Binary  & 34.81  & 78.13 \\
Expand  & 30.34  & 94.03 \\
Refresh & 5300.18 & 94.58 \\ \hline

\end{tabular}
\caption{ Comparison of per-epoch training time (seconds) and top-1 accuracy (\%) on an NVIDIA Tesla M40 GPU.}
\label{tab:compare_expansion}
\end{table}

We show the results when new domains are added to an IPDA and the model needs to efficiently accommodate them with a limited number of data samples. We show the classification performance on the skills in the \emph{New Skills} dataset while assuming we have access to the \emph{WEAK} dataset to pre-train our model for transfer learning. In the \texttt{Binary} setting, a domain-specific binary classifier is trained for each domain. \texttt{Expand} describes the case in which we incrementally train on top of an existing model. \texttt{Refresh} is the setting in which the model is fully retrained from scratch with the new data - which would be ideal in case there were no time constraints.

We record the average training time for each epoch and the performance is measured with top-1 classification accuracy over new skills. The experiment results can be found in Table~\ref{tab:compare_expansion}. Adapting a new skill is two orders of magnitude faster (30.34 seconds) than retraining the model (5300.18 seconds) while achieving 94.03\% accuracy which is comparable to 94.58\% accuracy of full retraining. The first two techniques can also be easily parallelized unlike the \texttt{Refresh} configuration. 
    
\paragraph{Behavior of Attention Mechanism}
Our expectation is that the model is able to learn to attend the relevant skills during the inference process. To study the behavior of the attention layer, we compute the top-N prediction accuracy based on the most relevant skills defined by the attention weights. In this experiment, we considered the subset of users who had enabled more than 20 domains to exclude trivial cases\footnote{Thus, the random prediction accuracy on enabled domains is less than 5\% and across the Full domain list is 0.066\%}. The results are shown in Table~\ref{tab:attention-based}. When the model attends to the entire set of 1500 (\textit{Full}), the top-5 prediction accuracy is 20.41\%, which indicates that a large number of skills can process the utterance, and thus it is highly likely to miss the correct one in the top-5 predictions. This ambiguity issue can be significantly improved by users' enabled domain lists as proved by the accuracies (\textit{Enabled}): 85.62\% for top-1, 96.15\% for top-3, and 98.06\% for top-5.\footnote{Visual inspection of the embeddings confirms that meaningful clusters are learned. We see clusters related to home automation, commerce, cooking, trivia etc, we show some examples in Appendix \ref{sec:viz_emb}.} Thus the attention mechanism can thus be viewed as an initial soft selection which is then followed by a fine-grained selection at the classification stage.

\begin{table}[!t]
\setlength\belowcaptionskip{-5pt}
\small
\centering
\begin{tabular}{cccc}
           & Top-1 & Top-3 & Top-5 \\ \hline
Full       &  6.17 & 14.30 & 20.41 \\
Enabled    & 85.62 & 96.15 & 98.06 \\ \hline

\end{tabular}
\caption{Top-N prediction accuracy (\%) on the full skill set (Full) and only enabled skills (Enabled).}
\label{tab:attention-based}
\end{table}

\paragraph{End-to-End User Evaluation}

All intermediate metrics on this task are proxies to a human customer's eventual evaluation. In order to assess the user experience, we need to measure its end-to-end performance. For a brief end-to-end system evaluation, 983 utterances from 283 domains were randomly sampled from the test set in the large-scale IPDA setting. 15 human judges (male=12, female=3) rated the system responses, 1 judge per utterance, on a 5-point Likert scale with 1 being \textit{Terrible} and 5 being \textit{Perfect}. The judgment score of 3 or above was taken as \textit{SUCCESS} and 2 or below as \textit{DEFECT}. The end-to-end \textit{SUCCESS} rate, thus user satisfaction, was shown to be 95.52\%. The discrepancy between this score and the score produced on MTurk dataset indicates that even in cases in which the model makes classification mistakes, some of these interpretations remain perceptually meaningful to humans. 


\section{Conclusions}
\label{sec:conclusions}
We have described a neural model architecture to address large-scale skill classification in an IPDA used by tens of millions of users every day. We have described how personalization features and an attention mechanism can be used for handling ambiguity between domains. We have also shown that the model can be extended efficiently and incrementally for new domains, saving multiple orders of magnitude in terms of training time. The model also addresses practical constraints of having a low memory footprint, low latency and being easily parallelized, all of which are important characteristics for real production systems.





\bibliographystyle{acl_natbib}
\bibliography{naaclhlt2018}

\begin{thebibliography}{43}
\expandafter\ifx\csname natexlab\endcsname\relax\def\natexlab#1{#1}\fi

\bibitem[{Bahdanau et~al.(2014)Bahdanau, Cho, and Bengio}]{bahdanau2014neural}
Dzmitry Bahdanau, Kyunghyun Cho, and Yoshua Bengio. 2014.
\newblock Neural machine translation by jointly learning to align and
  translate.
\newblock \emph{arXiv preprint arXiv:1409.0473}.

\bibitem[{Bapna et~al.(2017)Bapna, Tur, Hakkani-Tur, and Heck}]{Bapna2017}
Ankur Bapna, Gokhan Tur, Dilek Hakkani-Tur, and Larry Heck. 2017.
\newblock Towards zero shot frame semantic parsing for domain scaling.
\newblock In \emph{Interspeech 2017}.

\bibitem[{Bhargava et~al.(2013)Bhargava, Celikyilmaz, Hakkani-Tur, and
  Sarikaya}]{Bhargava2013EasyCI}
A.~Bhargava, Asli Celikyilmaz, Dilek~Z. Hakkani-Tur, and Ruhi Sarikaya. 2013.
\newblock Easy contextual intent prediction and slot detection.
\newblock \emph{IEEE International Conference on Acoustics, Speech and Signal
  Processing}, pages 8337--8341.

\bibitem[{Celikyilmaz et~al.(2016)Celikyilmaz, Sarikaya, Hakkani-T{\"u}r, Liu,
  Ramesh, and T{\"u}r}]{celikyilmaz2016new}
Asli Celikyilmaz, Ruhi Sarikaya, Dilek Hakkani-T{\"u}r, Xiaohu Liu, Nikhil
  Ramesh, and G{\"o}khan T{\"u}r. 2016.
\newblock A new pre-training method for training deep learning models with
  application to spoken language understanding.
\newblock In \emph{Interspeech}, pages 3255--3259.

\bibitem[{Chen et~al.(2016{\natexlab{a}})Chen, Hakkani-T{\"u}r, and
  He}]{chen2016zero}
Yun-Nung Chen, Dilek Hakkani-T{\"u}r, and Xiaodong He. 2016{\natexlab{a}}.
\newblock Zero-shot learning of intent embeddings for expansion by
  convolutional deep structured semantic models.
\newblock In \emph{Acoustics, Speech and Signal Processing (ICASSP), 2016 IEEE
  International Conference on}, pages 6045--6049.

\bibitem[{Chen et~al.(2016{\natexlab{b}})Chen, Hakkani-T{\"u}r, Tur, Gao, and
  Deng}]{chen2016end}
Yun-Nung Chen, Dilek Hakkani-T{\"u}r, Gokhan Tur, Jianfeng Gao, and Li~Deng.
  2016{\natexlab{b}}.
\newblock End-to-end memory networks with knowledge carryover for multi-turn
  spoken language understanding.
\newblock In \emph{Interspeech}.

\bibitem[{El-Kahky et~al.(2014)El-Kahky, Liu, Sarikaya, Tur, Hakkani-Tur, and
  Heck}]{el2014extending}
Ali El-Kahky, Xiaohu Liu, Ruhi Sarikaya, Gokhan Tur, Dilek Hakkani-Tur, and
  Larry Heck. 2014.
\newblock Extending domain coverage of language understanding systems via
  intent transfer between domains using knowledge graphs and search query click
  logs.
\newblock In \emph{IEEE International Conference on Acoustics, Speech and
  Signal Processing (ICASSP)}, pages 4067--4071. IEEE.

\bibitem[{Fan et~al.(2017)Fan, Monti, Mathias, and Dreyer}]{Fan2017TransferLF}
Xing Fan, Emilio Monti, Lambert Mathias, and Markus Dreyer. 2017.
\newblock Transfer learning for neural semantic parsing.
\newblock \emph{CoRR}, abs/1706.04326.

\bibitem[{Hochreiter and Schmidhuber(1997)}]{hochreiter1997long}
Sepp Hochreiter and J{\"u}rgen Schmidhuber. 1997.
\newblock Long short-term memory.
\newblock \emph{Neural computation}, 9(8):1735--1780.

\bibitem[{Hoffmann et~al.(2011)Hoffmann, Zhang, Ling, Zettlemoyer, and
  Weld}]{hoffmann2011knowledge}
Raphael Hoffmann, Congle Zhang, Xiao Ling, Luke Zettlemoyer, and Daniel~S Weld.
  2011.
\newblock Knowledge-based weak supervision for information extraction of
  overlapping relations.
\newblock In \emph{Proceedings of the 49th Annual Meeting of the Association
  for Computational Linguistics: Human Language Technologies-Volume 1}, pages
  541--550. Association for Computational Linguistics.

\bibitem[{Hori et~al.(2016)Hori, Hori, Watanabe, and Hershey}]{hori2016context}
Chiori Hori, Takaaki Hori, Shinji Watanabe, and John~R Hershey. 2016.
\newblock Context-sensitive and role-dependent spoken language understanding
  using bidirectional and attention lstms.
\newblock \emph{Interspeech}, pages 3236--3240.

\bibitem[{Jaech et~al.(2016)Jaech, Heck, and Ostendorf}]{jaech2016domain}
Aaron Jaech, Larry Heck, and Mari Ostendorf. 2016.
\newblock Domain adaptation of recurrent neural networks for natural language
  understanding.
\newblock In \emph{Interspeech}.

\bibitem[{Joulin et~al.(2016)Joulin, Grave, Bojanowski, and
  Mikolov}]{joulin2016bag}
Armand Joulin, Edouard Grave, Piotr Bojanowski, and Tomas Mikolov. 2016.
\newblock Bag of tricks for efficient text classification.
\newblock \emph{arXiv preprint arXiv:1607.01759}.

\bibitem[{Jozefowicz et~al.(2016)Jozefowicz, Vinyals, Schuster, Shazeer, and
  Wu}]{jozefowicz2016exploring}
Rafal Jozefowicz, Oriol Vinyals, Mike Schuster, Noam Shazeer, and Yonghui Wu.
  2016.
\newblock Exploring the limits of language modeling.
\newblock \emph{arXiv preprint arXiv:1602.02410}.

\bibitem[{Kernighan et~al.(1990)Kernighan, Church, and
  Gale}]{kernighan1990spelling}
Mark~D Kernighan, Kenneth~W Church, and William~A Gale. 1990.
\newblock A spelling correction program based on a noisy channel model.
\newblock In \emph{Proceedings of the 13th conference on Computational
  linguistics-Volume 2}, pages 205--210. Association for Computational
  Linguistics.

\bibitem[{Kim et~al.(2015{\natexlab{a}})Kim, Jeong, Stratos, and
  Sarikaya}]{kim2015weakly}
Young-Bum Kim, Minwoo Jeong, Karl Stratos, and Ruhi Sarikaya.
  2015{\natexlab{a}}.
\newblock Weakly supervised slot tagging with partially labeled sequences from
  web search click logs.
\newblock In \emph{Proceedings of the 2015 Conference of the North American
  Chapter of the Association for Computational Linguistics: Human Language
  Technologies}, pages 84--92.

\bibitem[{Kim et~al.(2017{\natexlab{a}})Kim, Lee, and
  Sarikaya}]{kim2017speaker}
Young-Bum Kim, Sungjin Lee, and Ruhi Sarikaya. 2017{\natexlab{a}}.
\newblock Speaker-sensitive dual memory networks for multi-turn slot tagging.
\newblock In \emph{Automatic Speech Recognition and Understanding Workshop
  (ASRU), 2017 IEEE}, pages 547--553. IEEE.

\bibitem[{Kim et~al.(2017{\natexlab{b}})Kim, Lee, and Stratos}]{kim2017onenet}
Young-Bum Kim, Sungjin Lee, and Karl Stratos. 2017{\natexlab{b}}.
\newblock Onenet: Joint domain, intent, slot prediction for spoken language
  understanding.
\newblock In \emph{Automatic Speech Recognition and Understanding Workshop
  (ASRU), 2017 IEEE}, pages 547--553. IEEE.

\bibitem[{Kim et~al.(2017{\natexlab{c}})Kim, Stratos, and Kim}]{kim2017advr}
Young-Bum Kim, Karl Stratos, and Dongchan Kim. 2017{\natexlab{c}}.
\newblock Adversarial adaptation of synthetic or stale data.
\newblock In \emph{Proceedings of the 55th Annual Meeting of the Association
  for Computational Linguistics}, pages 1297--1307. Association for
  Computational Linguistics.

\bibitem[{Kim et~al.(2017{\natexlab{d}})Kim, Stratos, and Kim}]{kim2017domain}
Young-Bum Kim, Karl Stratos, and Dongchan Kim. 2017{\natexlab{d}}.
\newblock Domain attention with an ensemble of experts.
\newblock In \emph{Annual Meeting of the Association for Computational
  Linguistics}.

\bibitem[{Kim et~al.(2015{\natexlab{b}})Kim, Stratos, and
  Sarikaya}]{kim2015pre}
Young-Bum Kim, Karl Stratos, and Ruhi Sarikaya. 2015{\natexlab{b}}.
\newblock Pre-training of hidden-unit crfs.
\newblock In \emph{Proceedings of the 53rd Annual Meeting of the Association
  for Computational Linguistics and the 7th International Joint Conference on
  Natural Language Processing}, volume~2, pages 192--198.

\bibitem[{Kim et~al.(2016)Kim, Stratos, and Sarikaya}]{kim2016frustratingly}
Young-Bum Kim, Karl Stratos, and Ruhi Sarikaya. 2016.
\newblock Frustratingly easy neural domain adaptation.
\newblock In \emph{Proceedings of COLING 2016, the 26th International
  Conference on Computational Linguistics: Technical Papers}, pages 387--396.

\bibitem[{Kim et~al.(2017{\natexlab{e}})Kim, Stratos, and
  Sarikaya}]{kim2017pre}
Young-Bum Kim, Karl Stratos, and Ruhi Sarikaya. 2017{\natexlab{e}}.
\newblock A framework for pre-training hidden-unit conditional random fields
  and its extension to long short term memory networks.
\newblock \emph{Computer Speech {\&} Language}, 46:311--326.

\bibitem[{Kim et~al.(2015{\natexlab{c}})Kim, Stratos, Sarikaya, and
  Jeong}]{kim2015new}
Young-Bum Kim, Karl Stratos, Ruhi Sarikaya, and Minwoo Jeong.
  2015{\natexlab{c}}.
\newblock New transfer learning techniques for disparate label sets.
\newblock In \emph{Proceedings of the 53rd Annual Meeting of the Association
  for Computational Linguistics and the 7th International Joint Conference on
  Natural Language Processing}, volume~1, pages 473--482.

\bibitem[{Klambauer et~al.(2017)Klambauer, Unterthiner, Mayr, and
  Hochreiter}]{Klambauer2017}
Gunter Klambauer, Thomas Unterthiner, Andreas Mayr, and Sepp Hochreiter. 2017.
\newblock Self-normalizing neural networks.
\newblock \emph{CoRR}, abs/1706.02515.

\bibitem[{Kubat et~al.(1997)Kubat, Matwin et~al.}]{kubat1997addressing}
Miroslav Kubat, Stan Matwin, et~al. 1997.
\newblock Addressing the curse of imbalanced training sets: one-sided
  selection.
\newblock In \emph{ICML}, volume~97, pages 179--186. Nashville, USA.

\bibitem[{Kumar et~al.(2017{\natexlab{a}})Kumar, Gupta, Chan, Tucker,
  Hoffmeister, and Dreyer}]{kumar2017ask}
Anjishnu Kumar, Arpit Gupta, Julian Chan, Sam Tucker, Bjorn Hoffmeister, and
  Markus Dreyer. 2017{\natexlab{a}}.
\newblock Just ask: Building an architecture for extensible self-service spoken
  language understanding.
\newblock \emph{arXiv preprint arXiv:1711.00549}.

\bibitem[{Kumar et~al.(2017{\natexlab{b}})Kumar, Muddireddy, Dreyer, and
  Hoffmeister}]{kumar2017zero}
Anjishnu Kumar, Pavankumar~Reddy Muddireddy, Markus Dreyer, and Bj{\"o}rn
  Hoffmeister. 2017{\natexlab{b}}.
\newblock Zero-shot learning across heterogeneous overlapping domains.
\newblock \emph{Proc. Interspeech 2017}, pages 2914--2918.

\bibitem[{Lample et~al.(2016)Lample, Ballesteros, Subramanian, Kawakami, and
  Dyer}]{lample2016neural}
Guillaume Lample, Miguel Ballesteros, Sandeep Subramanian, Kazuya Kawakami, and
  Chris Dyer. 2016.
\newblock Neural architectures for named entity recognition.
\newblock In \emph{Proceedings of NAACL-HLT}, pages 260--270.

\bibitem[{Liu and Lane(2016)}]{Liu+2016}
Bing Liu and Ian Lane. 2016.
\newblock Attention-based recurrent neural network models for joint intent
  detection and slot filling.
\newblock In \emph{Interspeech}, pages 685--689.

\bibitem[{Liu and Lane(2017)}]{liu2017multi}
Bing Liu and Ian Lane. 2017.
\newblock Multi-domain adversarial learning for slot filling in spoken language
  understanding.
\newblock In \emph{NIPS Workshop on Conversational AI}.

\bibitem[{van~der Maaten and Hinton(2008)}]{maaten2008tsne}
Laurens van~der Maaten and Geoffrey Hinton. 2008.
\newblock Visualizing high-dimensional data using t-sne.
\newblock \emph{Journal of Machine Learning Research}, 9:2579--2605.

\bibitem[{Mikolov et~al.(2013)Mikolov, Chen, Corrado, and
  Dean}]{mikolov2013efficient}
Tomas Mikolov, Kai Chen, Greg Corrado, and Jeffrey Dean. 2013.
\newblock Efficient estimation of word representations in vector space.
\newblock \emph{arXiv preprint arXiv:1301.3781}.

\bibitem[{Ratner et~al.(2016)Ratner, De~Sa, Wu, Selsam, and
  R{\'e}}]{ratner2016data}
Alexander~J Ratner, Christopher~M De~Sa, Sen Wu, Daniel Selsam, and Christopher
  R{\'e}. 2016.
\newblock Data programming: Creating large training sets, quickly.
\newblock In \emph{Advances in Neural Information Processing Systems}, pages
  3567--3575.

\bibitem[{Sarikaya(2017)}]{sarikaya2017SPM}
Ruhi Sarikaya. 2017.
\newblock The technology behind personal digital assistants: An overview of the
  system architecture and key components.
\newblock \emph{IEEE Signal Processing Magazine}, 34(1):67--81.

\bibitem[{Sarikaya et~al.(2016)Sarikaya, Crook, Marin, Jeong, Robichaud,
  Celikyilmaz, Kim, Rochette, Khan, Liu et~al.}]{sarikaya2016overview}
Ruhi Sarikaya, Paul~A Crook, Alex Marin, Minwoo Jeong, Jean-Philippe Robichaud,
  Asli Celikyilmaz, Young-Bum Kim, Alexandre Rochette, Omar~Zia Khan, Xiaohu
  Liu, et~al. 2016.
\newblock An overview of end-to-end language understanding and dialog
  management for personal digital assistants.
\newblock In \emph{Spoken Language Technology Workshop (SLT), 2016 IEEE}, pages
  391--397. IEEE.

\bibitem[{Seltzer and Droppo(2013)}]{seltzer2013multi}
Michael~L Seltzer and Jasha Droppo. 2013.
\newblock Multi-task learning in deep neural networks for improved phoneme
  recognition.
\newblock In \emph{Acoustics, Speech and Signal Processing (ICASSP), 2013 IEEE
  International Conference on}, pages 6965--6969. IEEE.

\bibitem[{Tur(2006)}]{tur2006multitask}
Gokhan Tur. 2006.
\newblock Multitask learning for spoken language understanding.
\newblock In \emph{Acoustics, Speech and Signal Processing, 2006. ICASSP 2006
  Proceedings. 2006 IEEE International Conference on}, volume~1, pages I--I.
  IEEE.

\bibitem[{Tur and de~Mori(2011)}]{Tur2011}
Gokhan Tur and Renato de~Mori. 2011.
\newblock \emph{Spoken Language Understanding: Systems for Extracting Semantic
  Information from Speech}.
\newblock New York, NY: John Wiley and Sons.

\bibitem[{Xiao and Cho(2016)}]{xiao2016efficient}
Yijun Xiao and Kyunghyun Cho. 2016.
\newblock Efficient character-level document classification by combining
  convolution and recurrent layers.
\newblock \emph{arXiv preprint arXiv:1602.00367}.

\bibitem[{Xu and Sarikaya(2014)}]{xu2014contextual}
Puyang Xu and Ruhi Sarikaya. 2014.
\newblock Contextual domain classification in spoken language understanding
  systems using recurrent neural network.
\newblock In \emph{Acoustics, Speech and Signal Processing (ICASSP), 2014 IEEE
  International Conference on}, pages 136--140. IEEE.

\bibitem[{Yang et~al.(2017)Yang, Salakhutdinov, and Cohen}]{yang2017transfer}
Zhilin Yang, Ruslan Salakhutdinov, and William~W Cohen. 2017.
\newblock Transfer learning for sequence tagging with hierarchical recurrent
  networks.
\newblock \emph{International Conference on Learning Representation (ICLR)}.

\bibitem[{Zhang and Wang(2016)}]{zhang2016joint}
Xiaodong Zhang and Houfeng Wang. 2016.
\newblock A joint model of intent determination and slot filling for spoken
  language understanding.
\newblock In \emph{International Joint Conference on Artificial Intelligence
  (IJCAI)}, pages 2993--2999.

\end{thebibliography}

\appendix

\newpage
\appendix
\section{Extended Analysis}
\label{sec:extended_analysis}
In this section we present a set of analyses performed on the performance of the new model.

\begin{figure*}[!h]
    \centering
    \includegraphics[width=1.0\textwidth]{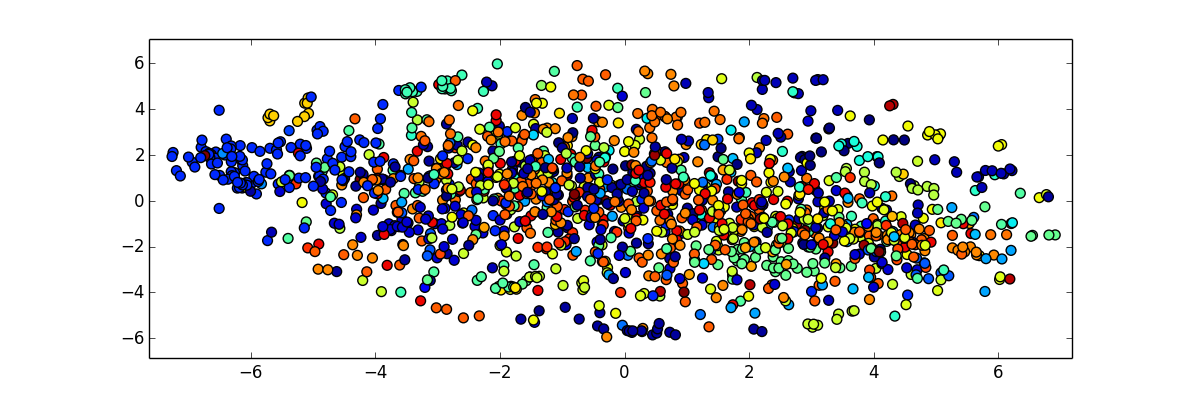}
    \caption{\small Embeddings of different domain categories visualized in 2D using TSNE \cite{maaten2008tsne}. Different colors represent different categories, for e.g. the large blue cluster on the left is Home Automation.}
    \label{fig:domain_embeddings}
\end{figure*}

\subsection{Errors by Category}

\begin{table}[!ht]
\centering
\begin{tabular}{cc}
Domain Category                 & Portion   \\ \hline
\texttt{HOME AUTOMATION}        & 22.6\%    \\
\texttt{SHOPPING}               & 11.8\%    \\
\texttt{CHITCHAT}               & 10.1\%    \\
\texttt{KNOWLEDGE}              &  9.3\%    \\
\texttt{COOKING}                &  6.2\%    \\ \hline
\end{tabular}
\caption{Error composition of different domains}
\label{tab:domain_err_comp}
\end{table}

We conducted detailed error analysis and the contributions of different domain categories to the errors is shown in Table~\ref{tab:domain_err_comp}. Domains belonging to the \texttt{HOME AUTOMATION} category\footnote{Domains in this category are designed to control smart devices at home, examples include \texttt{Harmony, Vivint, LIFX, etc}} are the largest contributor with 22.6\% of all errors. Domains belonging to this category have similar phrasing, for example, \textit{turn on/off} and \textit{start/stop/resume} patterns are all common for these domains. Furthermore, it is also quite common for customers to use multiple smart devices, making it more likely for different domains in this category to be enabled for a single customer.

\subsection{Visualizing Domain Embeddings}
\label{sec:viz_emb}
In this appendix we look more closely at some of the clusters of domains observed after using TSNE for dimensionality reduction.

We notice that many clusters are that are locally formed can be observed to have human interpretable semantic meaning. Some of these are visualized below. However there are still other clusters where the the relationships cannot be established as easily. An example of these is show in figure \ref{fig:mixed_cluster}. More detailed analysis will be necessary to understand why the model chooses to embed these domains close to each other. It could be that the effects of the patterns in the way users enable these domains dominate over the common semantic structures in language.

\begin{figure*}[ht]
    \centering
    \includegraphics[width=\textwidth]{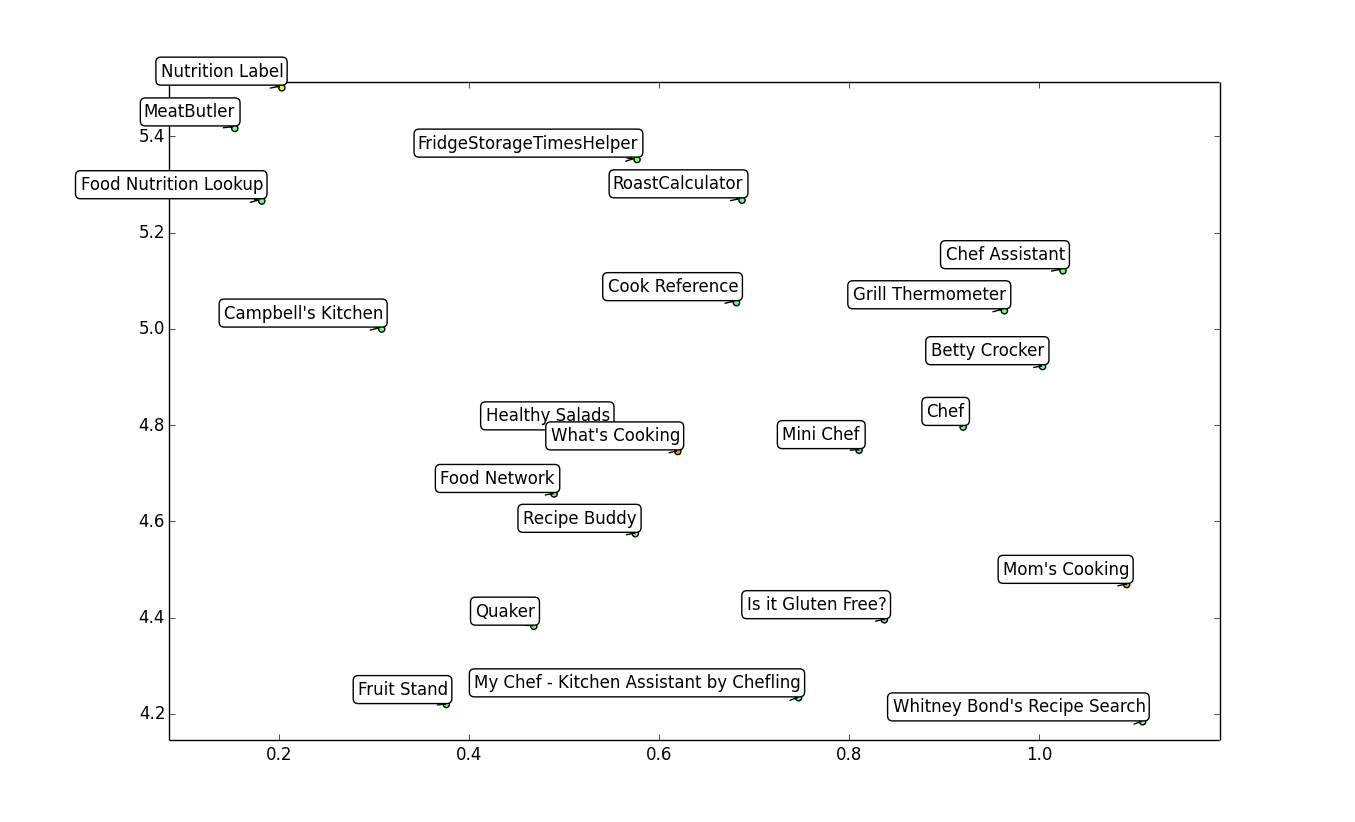}
    \caption{\small A cluster of domains related to cooking.}
    \label{fig:cooking_cluster}
\end{figure*}

\begin{figure*}[ht]
    \centering
    \includegraphics[width=\textwidth]{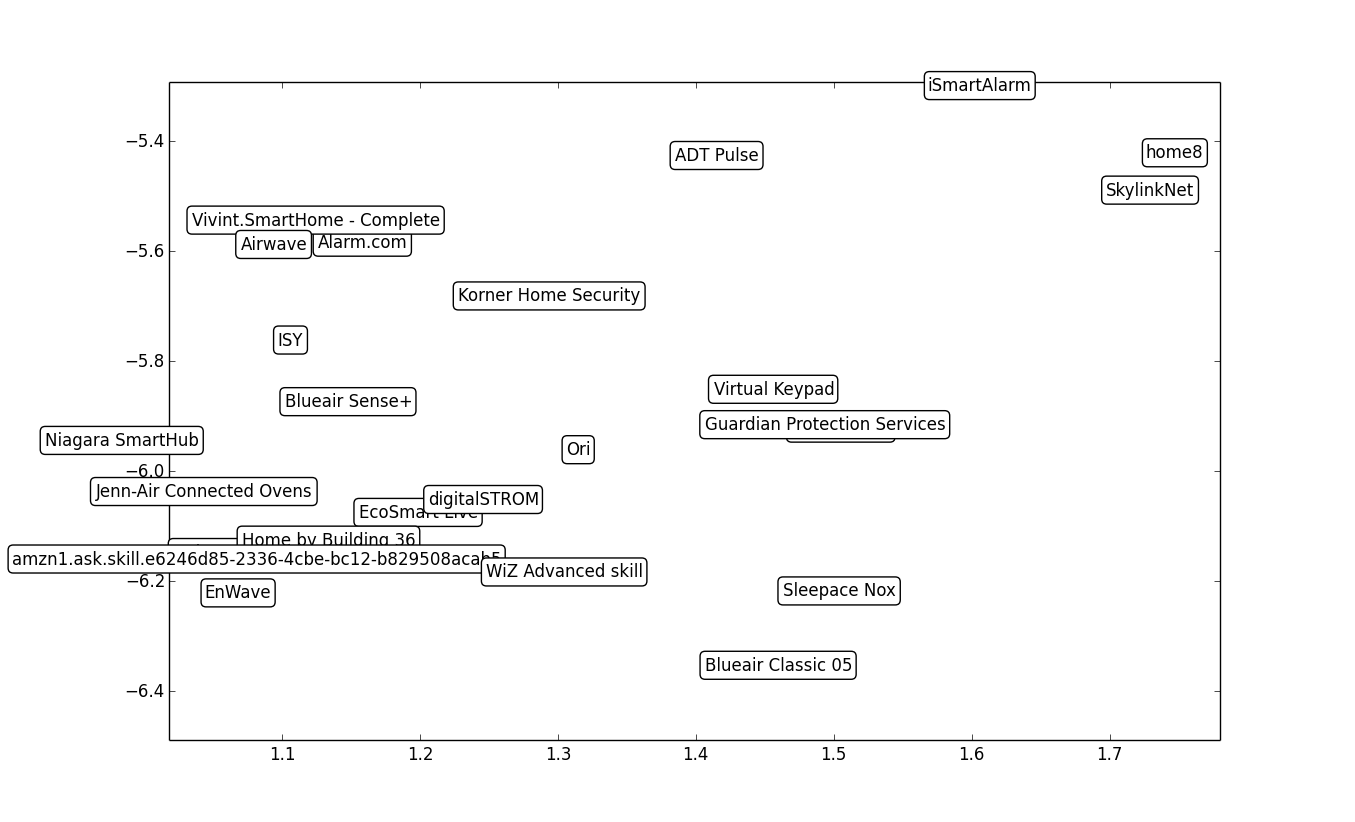}
    \caption{\small A large cluster of home automation domains.}
    \label{fig:home_automation_cluster}
\end{figure*}

\begin{figure*}[ht]
    \centering
    \includegraphics[width=\textwidth]{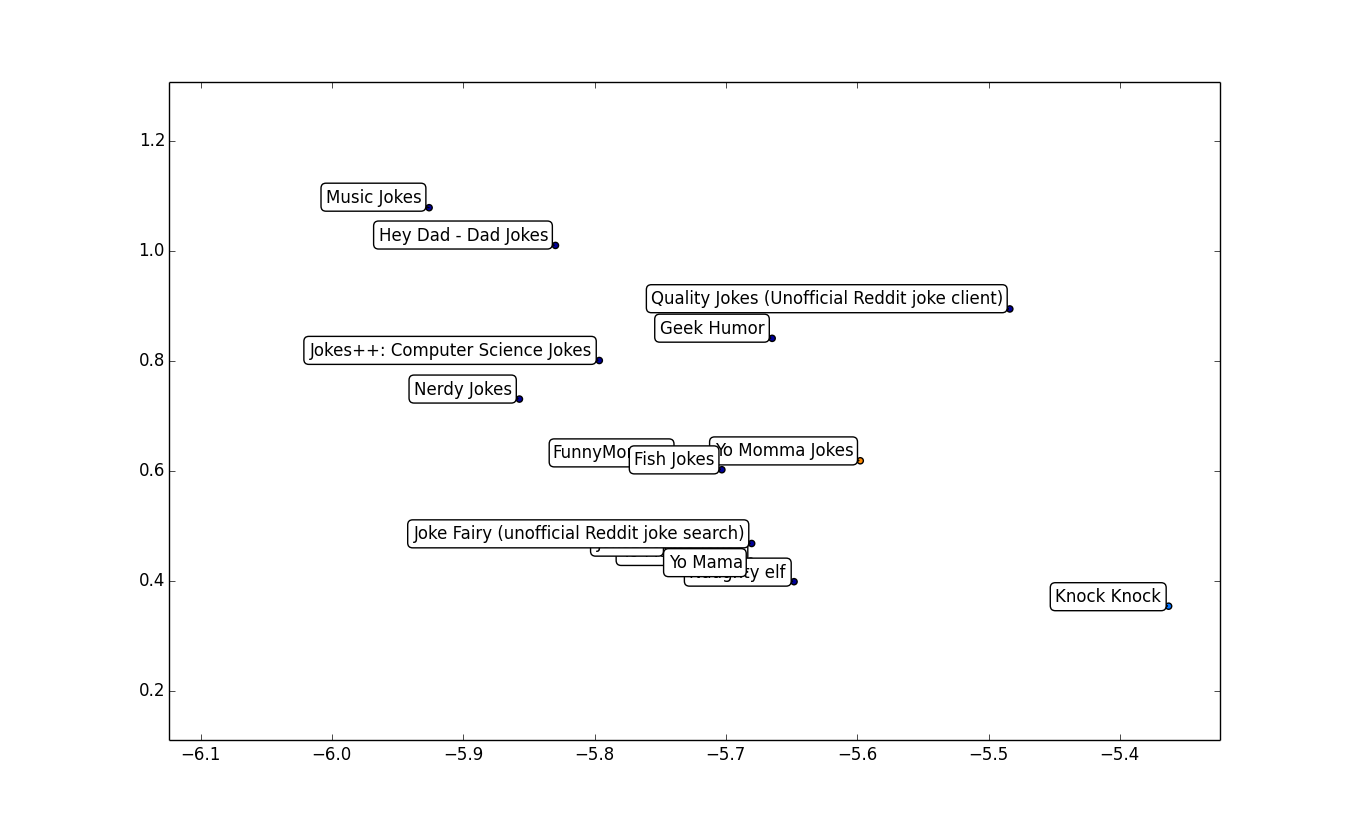}
    \caption{\small A cluster of domains.}
    \label{fig:joke_cluster}
\end{figure*}

\begin{figure*}[ht]
    \centering
    \includegraphics[width=\textwidth]{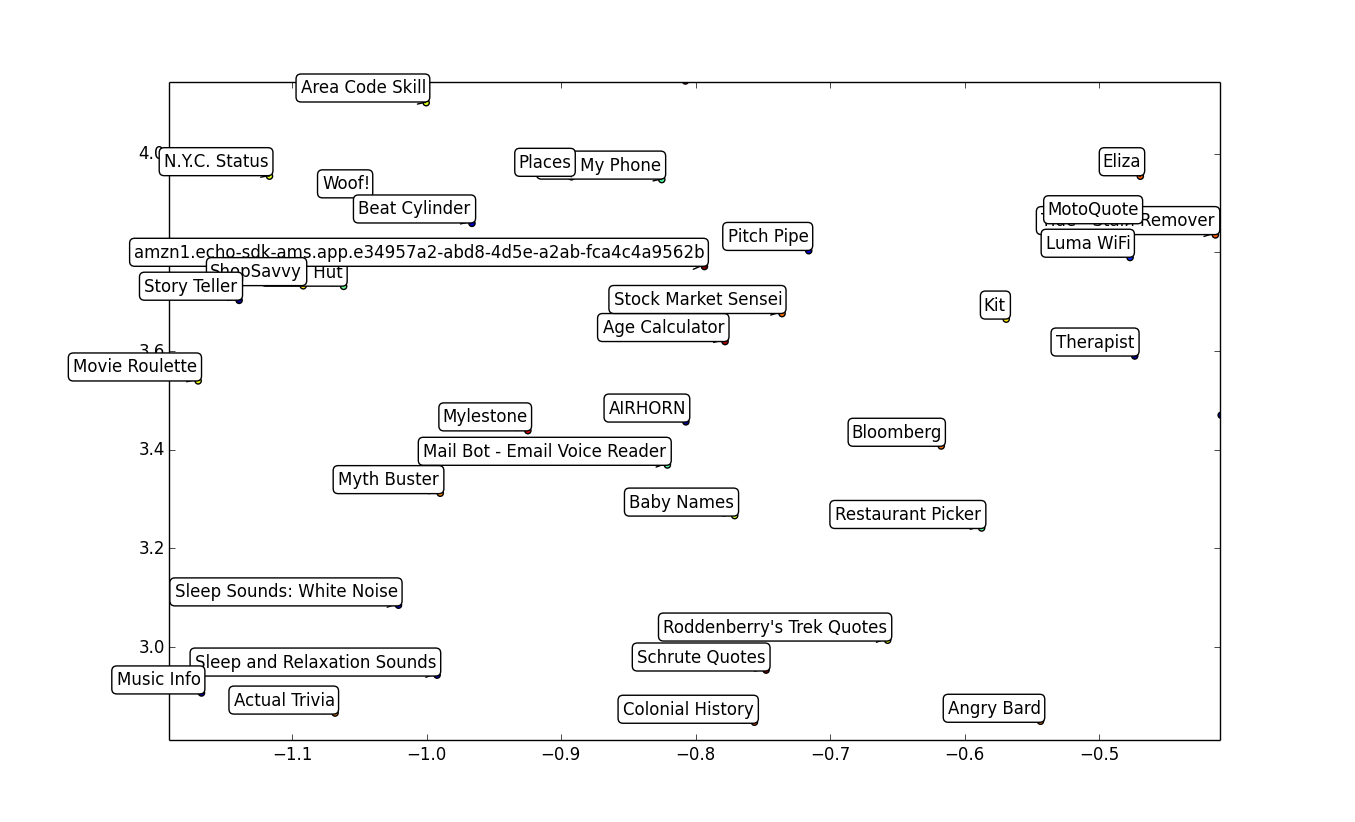}
    \caption{\small A mixed cluster with several different domain categories represented. The personalized attention mechanism is learned using the semantic content as well as personalization signals, so we hypothesize clusters like this may be capturing user tendencies to enable these domains in a correlated manner.}
    \label{fig:mixed_cluster}
\end{figure*}

\end{document}